\documentclass[final,twocolumn,3p]{elsarticle}

\usepackage{subcaption}
\usepackage{amssymb}
\usepackage{amsmath}
\usepackage{graphicx}
\usepackage{multirow}
\usepackage{booktabs}
\usepackage{xcolor}
\usepackage{colortbl}
\usepackage{arydshln}
\usepackage[colorlinks=True,urlcolor=magenta]{hyperref}
\usepackage{hhline}
\usepackage{soul}

\newcolumntype{s}{>{\columncolor{gray!10}}c}

\newcommand{\specialcell}[2][c]{\begin{tabular}[#1]{@{}c@{}}#2\end{tabular}}

\journal{Medical Image Analysis}

\biboptions{authoryear}

\begin{document}

\title{Anatomically-aware Uncertainty for Semi-supervised Image Segmentation}

\author[ets]{Sukesh Adiga V\corref{cor1}}
\author[ets]{Jose Dolz}
\author[ets]{Herve Lombaert}

\cortext[cor1]{Corresponding author: Sukesh Adiga V, \textbf{Email:} \textit{sukesh.adiga-vasudeva.1@ens.etsmtl.ca}. All authors are with Computer and Software Engineering Department, ETS Montreal, 1100 Notre Dame St. W., Montreal QC, H3C 1K3, Canada}
\address[ets]{ETS Montreal, Canada}

\begin{frontmatter}
\begin{abstract}

Semi-supervised learning relaxes the need of large pixel-wise labeled datasets for image segmentation by leveraging unlabeled data.
A prominent way to exploit unlabeled data is to regularize model predictions. Since the predictions of unlabeled data can be unreliable, uncertainty-aware schemes are typically employed to gradually learn from meaningful and reliable predictions. Uncertainty estimation methods, however, rely on multiple inferences from the model predictions that must be computed for each training step, which is computationally expensive. Moreover, these uncertainty maps capture pixel-wise disparities and do not consider global information. This work proposes a novel method to estimate segmentation uncertainty by leveraging global information from the segmentation masks. More precisely, an anatomically-aware representation is first learnt to model the available segmentation masks. The learnt representation thereupon maps the prediction of a new segmentation into an anatomically-plausible segmentation. The deviation from the plausible segmentation aids in estimating the underlying pixel-level uncertainty in order to further guide the segmentation network. The proposed method consequently estimates the uncertainty using a single inference from our representation, thereby reducing the total computation. We evaluate our method on two publicly available segmentation datasets of left atria in cardiac MRIs and of multiple organs in abdominal CTs. Our anatomically-aware method improves the segmentation accuracy over the state-of-the-art semi-supervised methods in terms of two commonly used evaluation metrics.

\end{abstract}

\end{frontmatter}

\section{Introduction}

Segmentation is a fundamental task in medical image analysis, where image pixels are associated with a target object, such as an organ, structure, or abnormal region. It is a vital pre-processing step in many clinical applications, notably in computer-assisted diagnosis, intervention assistance, treatment planning, and personalized medicine \citep{duncan2000medical,ayache201620th}. Recent segmentation methods based on deep learning techniques are driving progress under the full-supervision regime, often outperforming traditional methods \citep{litjens2017survey}. Such a regime, however, relies on a large amount of annotations, which is time-consuming. Delineating an image at a pixel-level is indeed challenging, especially in homogeneous or low-contrast regions, and requires prohibitive clinical expertise. The burden of image annotation motivates new learning strategies with limited supervision \citep{cheplygina2019not}.

\begin{figure*}[!ht]
\centering
\includegraphics[width=0.975\linewidth]{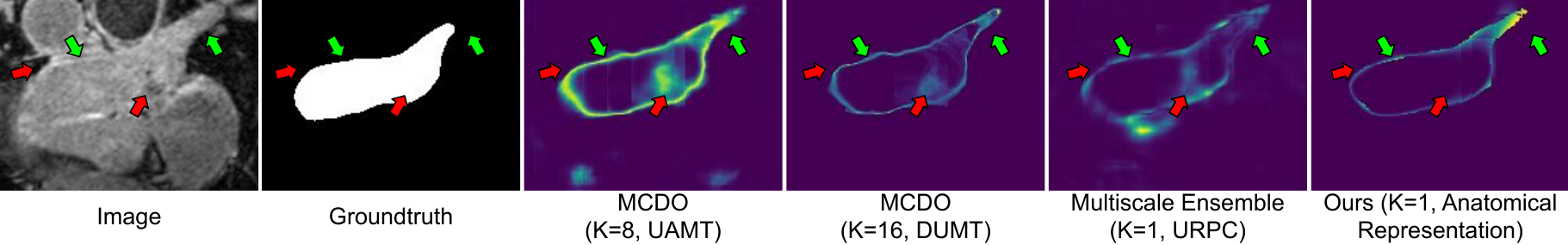}
\caption{\textbf{Uncertainty maps from different semi-supervision methods}. K denotes the number of inferences. Green arrows in regions of probable uncertainty due to unclear boundaries or annotator cut preference (such as in pulmonary veins cut in top right). Red arrows in regions of lower uncertainty as they depict high image gradients in uninformative clear boundary or inner foreground content.}
\label{fig:uncertainty_map}
\end{figure*}

Semi-supervised learning is an emerging strategy that alleviates annotation scarcity by leveraging unlabeled data with a small set of labeled data. Current semi-supervised segmentation methods typically utilize unlabeled data either in the form of pseudo labels \citep{bai2017semi,zheng2020cartilage} or in a regularization term \citep{nie2018asdnet,cui2019semi,peng2020deep}. The former strategies augment the original labeled dataset with unlabeled data alongside its corresponding model predictions, commonly referred to as pseudo labels. Later techniques incorporate unlabeled data into the training process by constraining predictions with a regularizer term. Training these semi-supervised approaches typically involves a supervised loss associated with labeled data and an unsupervised loss associated with unlabeled data.

Among regularization techniques, consistency-based approaches \citep{laine2016temporal,tarvainen2017mean} are often used in semi-supervision due to simple ways to leverage unlabeled data. Their approach encourages two or more segmentation predictions to be consistent under different perturbations of the input data \citep{cui2019semi,bortsova2019semi,li2020transformation}. However, the segmentation predictions can be unreliable and noisy for unlabeled data since its annotations are unavailable. To alleviate this issue, uncertainty-aware regularization methods \citep{yu2019uncertainty,sedai2019uncertainty} have been proposed to gradually add reliable target regions in predictions. Although these methods perform well in low-labeled data regime, their high computation and complex training techniques remain a limiting factor to broader applications. For instance, the pixel-level uncertainty approximation with Monte-Carlo Dropout (MCDO) \citep{gal2016dropout} or ensembling \citep{lakshminarayanan2017simple} requires multiple predictions per image, thereby increasing the computation of each training step. Moreover, these approaches do not consider global information to estimate uncertainty. The resulting uncertainty maps capture pixel-wise disparity, most likely around boundaries \citep{kendall2017bayesian}. However, high gradient regions near anatomical boundaries or inner content of anatomical structures should have a certain labeling mask. For instance, Fig.~\ref{fig:uncertainty_map} shows uncertainty captured by MCDO mostly over boundaries, while regions with high gradients (red arrows) could indicate certain boundaries or anatomical details with certainty. Probable uncertainty may lie in areas of low image gradients. For instance, anatomical boundaries may be unclear due to imaging or even non-existent in case of an arbitrary cut from an annotator (green arrows), as illustrated in the pulmonary veins in Fig.~\ref{fig:uncertainty_map}. Existing methods could benefit from capturing informative uncertainty in images beyond highlighting high image gradients or all over boundaries.

The global information of the anatomical regions is one promising direction to provide cues about informative uncertainty in images. Our approach will, therefore, exploit and capture global anatomical information by leveraging available masks to approximate segmentation uncertainty. Our main idea is to learn an anatomically-aware representation from a training set of segmentation masks. The learnt representation maps incorrect model predictions onto an anatomically-plausible segmentations. The plausible segmentation is subsequently used to estimate the uncertainty maps and further guide training of the segmentation network. We hypothesize that the proposed uncertainty estimates are more robust and computationally less expensive than deriving them from a standard entropy variance-based method, which requires multiple inferences for each training step.

\paragraph{\bf Our contributions}
We propose a novel approach to estimate the uncertainty maps from an anatomically-aware representation of the segmentation masks, in order to guide the training of a semi-supervised segmentation model. More precisely, we innovate semi-supervised segmentation with uncertainty-based training by integrating a pre-trained denoising autoencoder (DAE) into the training of our segmentation network to: (i) map the inaccurate model predictions to plausible segmentation masks and (ii) estimate new uncertainty maps that guide the training of our segmentation model. As we approximate the uncertainty based on the difference between predicted segmentation and its DAE reconstruction learned from the segmentation mask, it can better integrate anatomical information. In contrast to most uncertainty-based approaches, estimating the uncertainty map requires a single inference from the DAE model, thereby reducing computational complexity. Our method is extensively evaluated on two medical imaging datasets: the 2018 Atrial segmentation challenge dataset \citep{xiong2021global} and the 2021 Abdominal organ segmentation dataset \citep{ma2022fast}. Results demonstrate the superiority of our approach over the state-of-the-art methods in semi-supervised segmentation.

A preliminary version of this work has been published in MICCAI 2022 \citep{adiga2022leveraging}. This work includes a comprehensive literature review, extensive experiments, and a thorough discussion. The additional contributions in this manuscript are summarized as follows: (i)~an additional multi-class abdominal segmentation dataset is evaluated for all our experiments, including ablation studies; (ii)~the impact of various design choices made in the anatomically-aware representation prior (DAE) module are studied; (iii)~a qualitative comparative analysis of uncertainty for different methods and their computation time are provided; (iv)~additional related baseline that use a Monte-Carlo Dropout-based uncertainty estimation is provided for comparison \citep{wang2020double}; (v)~the introduction and motivation of our approach are significantly extended with illustrations of our uncertainty maps; (vi)~our literature review is expanded with recent uncertainty-aware as well as anatomically-plausible segmentation methods.

\subsection{Related Work}

\paragraph{Semi-Supervised Segmentation}
Semi-supervised learning (SSL) is an established approach in the literature under the paradigm of learning with limited supervision \citep{jiao2022learning}. A wide range of SSL strategies have been explored for segmentation, such as self-training \citep{bai2017semi,zheng2020cartilage}, entropy minimization \citep{grandvalet2004semi,wu2021semi}, consistency regularization \citep{cui2019semi,bortsova2019semi}, co-training \citep{peng2020deep,xia20203d} or adversarial learning \citep{nie2018asdnet,chaitanya2019semi}. For instance, self-training methods \citep{bai2017semi,zheng2020cartilage} typically employ pseudo-labels on unlabeled data to train models in an iterative way. However, potential labeling mistakes in the pseudo labels can quickly propagate during training, causing undesired segmentation outcomes. Entropy minimization strategies \citep{wu2021semi} circumvent such issues by enforcing a high confidence in predictions but can also easily lead to trivial solutions if additional priors are not used. Co-training approaches \citep{peng2020deep,xia20203d} avoid iterations but at the cost of simultaneously training two or more networks with multi-view images. Adversarial methods \citep{nie2018asdnet,chaitanya2019semi} encourage the predictions of unlabeled images to be closer to those of the labeled images, however, they remain challenging in terms of convergence \citep{salimans2016improved}. Among the existing SSL strategies, consistency regularization-based methods \citep{laine2016temporal,tarvainen2017mean} are popular due to their simple assumption that predictions should not change significantly under different realistic data perturbations. This notion is formulated as a consistency regularization term in the loss function, which encourages predictions to be consistent between data and its perturbed version \citep{cui2019semi,bortsova2019semi,li2020transformation}. Similarly, our method leverages unlabeled data with a consistency regularizer.

\paragraph{Uncertainty-based methods}
The uncertainty estimation approaches often employ Bayesian neural networks \citep{neal2012bayesian}, however, their training process poses significant computational challenges. Recent deep learning methods address this limitation by approximating uncertainty through the generation of multiple samples \citep{abdar2021review}. For instance, Monte-Carlo Dropout (MCDO) \citep{gal2016dropout} performs several forward passes through the same model with dropout enabled at test time to generate multiple samples for the same input. Whereas a deep ensemble \citep{lakshminarayanan2017simple} trains a set of independent models to generate multiple samples. These approaches, however, tackle the problem of approximating \textit{epistemic} uncertainty associated with the model output but not the \textit{aleatoric} uncertainty associated with the model input \citep{kendall2017uncertainties}. A set of recent methods models the \textit{aleatoric} uncertainty by using intra-/inter-annotation variability as a proxy to the underlying input uncertainties \citep{kohl2018probabilistic,baumgartner2019phiseg,monteiro2020stochastic}. All of the aforementioned methods have been shown to produce reliable uncertainty estimations in fully-supervised segmentation \citep{mehta2022qu,camarasa2021quantitative}.

In the context of semi-supervised segmentation, the uncertainty in the prediction is widely used within the optimization process \citep{yu2019uncertainty,wang2020double,wang2021tripled}. In particular, the uncertainty information assists the segmentation models by providing reliable target regions on unlabeled data during each training step. For instance, \cite{yu2019uncertainty} first approximates an uncertainty map using a predictive entropy of several predictions under data and model perturbations. The generated uncertainty map is later used to gradually add the reliable target regions in the consistency loss term. This idea was further extended to integrate uncertainty on a feature-level \citep{wang2020double} and multiple prediction branches \citep{wang2022semi}. The uncertainty estimation in these approaches commonly use MCDO \citep{gal2016dropout} or ensembling \citep{lakshminarayanan2017simple}, which inherently relies on multiple predictions per image. In addition to being computationally expensive, estimating such entropy-based uncertainty is suboptimal in a multi-class scenario since it disregards inter-class overlaps  \citep{van2022pitfalls}. More recently, multi-scale \citep{luo2022semi} or multi-decoder \citep{wu2022mutual} approaches have been proposed to overcome the expensive computation of uncertainty using multiple predictions in a single forward pass. Nevertheless, these methods failed to capture the actual uncertainty regions. In contrast to existing strategies, our method leverages an anatomically-aware representation from the available annotations to estimate the uncertainty in a single inference step. This strategy leads to a lower computational complexity and an improved computational efficiency.

\paragraph{Towards anatomically-plausible segmentations}
Recent approaches incorporate anatomically-aware priors in a segmentation network \citep{oktay2017anatomically,ravishankar2017learning,painchaud2020cardiac} by learning the variability of structures in a medical imaging dataset. For instance, \cite{oktay2017anatomically} first learn an anatomically-aware representation with an autoencoder-based architecture using segmentation masks. This representation is later utilized to map a prediction into an anatomically-plausible space. These methods use the encoder of the representation as a global shape regularizer that enforces the model predictions to follow the ground truth distribution. The anatomically-aware representation can also map an erroneous mask into an anatomically-plausible segmentation. Such mapping is subsequently used to correct the segmentation predictions as a post-processing step \citep{larrazabal2020post,painchaud2020cardiac} or improve the segmentation on unseen test images \citep{karani2021test}. In order to encode the masks in the anatomically-aware representation, a substantial amount of annotations are used either from the given dataset \citep{larrazabal2020post,painchaud2020cardiac} or the source domain dataset \citep{karani2021test}. The anatomically-aware representation is alternately substituted with a probabilistic atlas to enforce the priors \citep{zheng2019semi,huang2022mtl}, which requires an aligned dataset. For instance, \cite{dalca2018anatomical} learns an anatomically-aware representation on aligned labelings and subsequently uses it for unsupervised segmentation on aligned images. In contrast to these approaches, our method leverages an anatomically-aware representation in a low-data regime with the goal of obtaining uncertainty maps in order to guide the segmentation network during the training process.

\begin{figure*}[!t]
\centering
\includegraphics[width=0.975\linewidth]{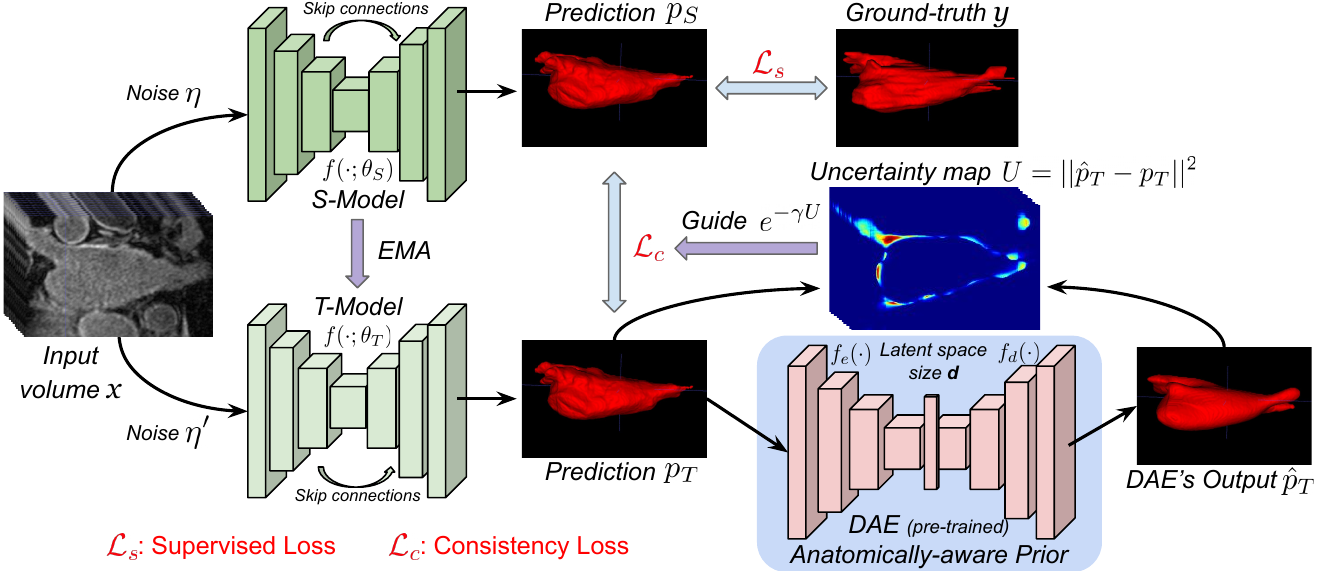}
\caption{\textbf{Overview of our uncertainty estimation from anatomically-aware representation for semi-supervised segmentation.} A pre-trained anatomically-aware representation (i.e., a DAE) module is integrated into the training of the mean teacher model, which maps the teacher prediction $p_T$ into a plausible segmentation $\hat{p}_T$. The uncertainty map ($U$) is subsequently estimated with the output of the teacher and the DAE model in order to further guide the student model. }
\label{fig:anatomical_rep_arch}
\end{figure*}

\section{Method}
\label{sec:methods}

An overview of the proposed anatomically-aware uncertainty estimation for semi-supervised segmentation is shown in Fig~\ref{fig:anatomical_rep_arch}. The main idea is to exploit an anatomically-aware representation that maps the segmentation prediction into a plausible mask. The reconstructed segmentation will be indicative in estimating an uncertainty map, which later is used to guide the segmentation training. The following subsections describe the semi-supervised setting, anatomically-aware representation and uncertainty estimation process.

\subsection{Preliminaries}
The standard semi-supervised learning consists of $N$ labeled and $M$ unlabeled data in the training set, where $N \ll M$. Let  $D_L = \{(x_i, y_i)\}_{i=1}^N$ and $D_U = \{(x_i)\}_{i=(N+1)}^{(N+M)}$ denote the labeled and unlabeled sets, where an input volume is represented as $x_i \in \mathbb{R}^{H \times W \times D}$ and its corresponding segmentation mask is $y_i \in \{0,1,...,C\}^{H \times W \times D}$, with $C$ being the number of classes. The objective is to train a segmentation network with a combination of supervised loss $\mathcal{L}_s$ and unsupervised loss $\mathcal{L}_u$ using labeled and unlabeled data, i.e., $\mathcal{L} = \mathcal{L}_s + \lambda \mathcal{L}_u$, where $\lambda$ controls the weight of unsupervised loss.

\subsection{Mean Teacher Formulation}
Following current literature \citep{yu2019uncertainty}, we adopt the common mean teacher approach \citep{tarvainen2017mean} for training a segmentation network. It consists of a student ($S$) and a teacher ($T$) model, both having the same segmentation architecture. The overall objective function is defined as follows:
\begin{multline}
\label{eq:ssl}
\mathcal{L} = \underset{\theta_S}{\text{min}} \sum_{i=1}^N \mathcal{L}_s(f(x_i; \theta_S), y_i) + \\
\lambda_c \sum_{i=1}^{N+M} \mathcal{L}_c(f(x_i, \eta; \theta_S), f(x_i, \eta{'}; \theta_T)),
\end{multline}
where $f(\cdot)$ denotes the segmentation network, and $\theta_S$ and $\theta_T$ are the learnable weights of the student and teacher models. The supervised loss $\mathcal{L}_s$ measures the segmentation quality on the labeled data, whereas the unsupervised consistency loss ($\mathcal{L}_c = \mathcal{L}_u$) measures the prediction consistency between the student and the teacher models for the same input volume $x_i$ under different perturbations ($\eta$ and $\eta{'}$). The balance between the supervised and unsupervised loss is controlled by a ramp-up weighting coefficient $\lambda_c$, which is defined as
\begin{equation}
\label{eq:lambda}
    \lambda_c = \beta * e^{-r (1-\frac{t}{t_{max}})^2},
\end{equation}
where $\beta$ is a consistency weight, $r$ controls the rate of ramp-up, $t$ and $t_{max}$ denote the current and maximum training steps. For training, the student model parameters ($\theta_S$) are optimized with stochastic gradient descent (SGD), whereas the teacher model parameters ($\theta_T$) are updated using an exponential moving average (EMA) at each training step $t$. The EMA is defined as 
\begin{equation}
\label{eq:EMA}
    \theta_T^t = \alpha \theta_T^{t-1} + (1-\alpha) \theta_S^t,
\end{equation}
where $\alpha$ is the smoothing coefficient of EMA that controls the update rate.

\subsection{Anatomically-aware Uncertainty Approach}
The reliability of the model prediction on the unlabeled dataset plays an essential role in the consistency loss. An uncertainty-aware scheme can assist this loss by providing reliable target regions. The existing approaches \citep{yu2019uncertainty,wang2020double} estimate uncertainty at a pixel-level, which fails to consider global information within the dataset. To address this limitation, our approach learns an anatomically-aware representation prior in order to capture global information. The measurable deviations from this prior provide informative cues about the uncertainty of the segmentation mask. The following subsections elaborate on our anatomically-aware uncertainty method. 

\subsubsection{Anatomically-aware Representation Prior}
\label{sec:prior}
Incorporating anatomically-aware prior in deep segmentation models is not obvious. One of the reasons is that, in order to integrate such prior knowledge during training, one needs to augment the learning objective with a differentiable term, which is not trivial. To circumvent these difficulties, a simpler solution is to resort to an autoencoder trained with segmentation masks, which maps the predictions into anatomically-plausible segmentation. This strategy has been adopted for fully-supervised learning as a global regularizer during training in \citep{oktay2017anatomically} and as a post-processing step in \citep{larrazabal2020post} to correct the segmentation predictions.
Motivated by this concept, we encode the available segmentation masks in a non-linear latent space of a denoising autoencoder (DAE) \citep{vincent2010stacked} to learn an anatomically-aware representation prior. This learnt representation captures the global information from the segmentation masks such that it maps an inaccurate prediction into a plausible segmentation.

The DAE model consists of an encoder $f_e(\cdot)$ and a decoder $f_d(\cdot)$ with a $d$-dimensional latent space as shown in the Fig.~\ref{fig:anatomical_rep_arch}. The DAE is trained to reconstruct the clean label $y_i$ from its corrupted version $\tilde{y}_i$, which can be achieved with a mean squared error loss: $\frac{1}{H \times W \times D}\sum_v ||f_d(f_e(\tilde{y}_{i,v})) - y_{i,v}||^2$, where $v$ is a voxel. Additionally, the dice loss is added to handle the class imbalance between foreground and background in the labels.

\subsubsection{Anatomically-aware Uncertainty}
\label{ssec:uncertainty_anatomical_prior}
The role of the uncertainty is to gradually update the student model with reliable target regions from the teacher model predictions. Our proposed method estimates the uncertainty directly from the anatomically-aware representation network $f_d(f_e(\cdot))$, requiring only one inference step. First, we map the segmentation prediction from the teacher model ($p_{T_i}$) with a DAE model to produce a plausible segmentation $\hat{p}_{T_i}$ = $f_d(f_e(p_{T_i}))$. We subsequently estimate the uncertainty as the pixel-wise difference between the DAE output and the prediction, which is given as:
\begin{equation}
\label{eq:uncertainty_formulation}
U_i = ||\hat{p}_{T_i} - p_{T_i}||^2.
\end{equation}

Note that the uncertainty formulation is related to the conventional sample variance-based uncertainty estimation. Specifically, for a given input, $x_i$, and its corresponding multiple model predictions, $p_{i_s}$, the sample variance estimation is defined as follows:
\begin{equation*}
var(p_i) =  \frac{1}{S-1} \sum_{s=1}^{S} (p_{i_s} - \bar{p}_i)^2,
\end{equation*}
where $\bar{p}_i$ represents the sample mean and is defined as $\bar{p}_i = \frac{1}{S} \sum_{s=1}^{S} (p_{i_s})$. The parameter $S$ denotes the number of prediction samples. When $S$ is set to 2, the sample mean $\bar{p}_i$ reduces to $\frac{p_{i_1} + p_{i_2}}{2}$, resulting in the variance estimation taking the form of:
\begin{align*}
var(p_i) &=  (p_{i_1} - \frac{p_{i_1} + p_{i_2}}{2})^2 + (p_{i_2}-\frac{p_{i_1} + p_{i_2}}{2})^2 , \\
&=  (\frac{p_{i_1} - p_{i_2}}{2})^2 + (\frac{p_{i_2} - p_{i_1}}{2})^2 , \\
var(p_i) &= \frac{1}{2} (p_{i_1} - p_{i_2})^2 .
\end{align*}

The above equation is equivalent to our uncertainty formulation in Eq.~\ref{eq:uncertainty_formulation}, where two samples are drawn from the output of the teacher model and the DAE model. 

The resulting uncertainty maps from Eq.~\ref{eq:uncertainty_formulation} are subsequently used to obtain the reliable target regions as follows: $e^{-\gamma U_i}$, similarly to \citep{luo2022semi}, where $\gamma$ is an uncertainty weighting factor empirically set to 1. The reliable targets are finally combined in a consistency loss as:

\begin{equation}
\label{eq:consistency_loss}
\mathcal{L}_c(p_{S_i}, p_{T_i}) = \frac{\sum_v e^{-\gamma U_{i,v}} ||p_{S_{i,v}} - p_{T_{i,v}}||^2}{\sum_v e^{-\gamma U_{i,v}}},
\end{equation}
where $v$ is a voxel. Note that the consistency loss $\mathcal{L}_c$ will be equivalent to a standard mean teacher method \citep{tarvainen2017mean} when $\gamma=0$. Overall, we jointly optimize the consistency loss $\mathcal{L}_c$ and supervised loss $\mathcal{L}_s$ as learning objectives, where $\mathcal{L}_s$ is a combination of cross-entropy and dice losses.

\section{Experiments}
\subsection{Datasets}
The performance of our method is validated on two publicly available benchmarks: (a) the left atrium (LA) binary segmentation dataset from the 2018 atrial challenge \citep{xiong2021global}, and (b) the abdominal multi-organ segmentation dataset from the FLARE challenge \citep{ma2022fast}. 

\paragraph{(a) LA dataset}
It consists of 100 3D late gadolinium-enhanced magnetic resonance imaging (LGE-MRI) scans and corresponding LA segmentation masks. These scans have an isotropic resolution of $0.625~mm^3$ and are center cropped at the heart region. The dataset is split into 80 for training and the remaining 20 for testing as in the literature \citep{yu2019uncertainty,li2020shape,wang2020double,luo2021semi}.

\paragraph{(b) FLARE dataset}
This dataset consists of 361 CT scans of the abdominal region and corresponding segmentation masks of four organs, namely liver, kidney, spleen, and pancreas. These scans are collected from multiple medical centers, having varying resolutions. Each image is first resampled to a uniform resolution of $2 \times 2 \times 2.5~mm^3$ and then normalized by clipping the intensity values outside $[0.5, 0.95]$ percentile range. For all our experiments, we use a fixed dataset split of 260 for training, 26 for validation, and the remaining 75 for testing. 

\subsection{Implementation and Training details}
\label{implementation_and_training_details}
To validate our proposed method, we employ a V-Net \citep{milletari2016v} as a backbone architecture for the segmentation networks, as followed in earlier work \citep{yu2019uncertainty,wang2020double,luo2021semi}. Our anatomically-aware representation prior module (i.e., a DAE) follows a similar architecture as V-Net but without skip connections. Such design effectively makes it an autoencoder-style architecture, which is also comparable to prior work \citep{oktay2017anatomically,larrazabal2020post}. To encode the segmentation mask in a latent space, a dense layer of $d$-dimension is added at the bottleneck layer of the DAE module as shown in Fig.~\ref{fig:anatomical_rep_arch}. For training, the student model uses a SGD optimizer with an initial learning rate ($lr$) of 0.1 and a momentum of 0.9 with a cosine annealing decaying \citep{loshchilov2016sgdr}. The teacher weights (in Eq.~\ref{eq:EMA}) are updated by an EMA with a rate of $\alpha=0.99$ \citep{tarvainen2017mean}. The DAE model is also trained using a SGD optimizer with an initial $lr=0.1$, a momentum of 0.9, and decaying the $lr$ by a factor of 2 every 5000 iterations. Following the literature \citep{yu2019uncertainty,luo2022semi}, the consistency weight $\beta$ and ramp-up factor $r$ in Eq.~\ref{eq:lambda} are set to 0.1 and 5, respectively. Inputs to both segmentation and DAE networks are randomly cropped to a size of $112 \times 112 \times 80$ and $144 \times 144 \times 96$ for LA and FLARE datasets, respectively. We employ online standard data augmentation techniques such as random flipping and rotation. In addition, input labels to the DAE are corrupted with a random swapping of pixels around class boundaries, morphological operations (erosion and dilation), resizing, and adding/removing basic shapes \citep{van2014scikit}. The latent space of the DAE is injected with a small noise drawn from a Gaussian distribution to explore different sets of plausible segmentation during training of the segmentation network. The training set is partitioned into $N$ labeled and $M$ unlabeled splits, which are fixed across all experiments. The batch size is set to 4 in both networks. Input batch for the segmentation network uses two labeled and unlabeled data. During the inference phase, the segmentation predictions are generated using the sliding window strategy. For the cardiac dataset (LA), following the literature \citep{yu2019uncertainty,li2020shape,luo2021semi}, the final model is evaluated at the last training iteration (i.e., 6000), whereas the best validation model is selected in the case of the abdominal dataset (FLARE). All our experiments were run on an NVIDIA RTX A6000 GPU with PyTorch 1.8.0. The implementation of our work is available at: \url{https://github.com/adigasu/Anatomically-aware_Uncertainty_for_Semi-supervised_Segmentation}.

\subsection{Evaluation Metrics}
We employ common Dice Score Coefficient (DSC) and 95\% Hausdorff Distance (HD) evaluation metrics to assess quantitative segmentation performance. The DSC score evaluates the degree of overlap between ground truth and prediction regions. In contrast, the HD score measures the distance between ground truth and predicted segmentation boundaries. For a fair comparison, all experiments are run three times with a fixed set of seeds on the same machine, and their average results are reported.

\section{Results}

\begin{table*}[!ht]
\footnotesize
\centering
\addtolength{\tabcolsep}{1pt}
\caption{\textbf{Segmentation results on the LA test set for the 10\% and 20\% annotation settings}. Uncertainty-based methods with $K$ inferences per training step are grouped at the bottom of each section, while $K$ = - indicates non-uncertainty-based methods. Ours achieves the best Dice (DSC) and Hausdorff (HD) scores in both annotation scenarios. The best and second-best results are highlighted in bold and underlined, whereas the statistical significance between the top two results is denoted in $^*$. The number of labeled and unlabeled data indicated with $N$ and $M$, respectively.}
\begin{tabular}{c | l | c | c c}
\rowcolor{gray!10} \bf $N$/$M$  & \bf Methods         & \bf \#$K$ &  \bf DSC (\%) $\uparrow$ & \bf HD (mm)  $\downarrow$ \\
\toprule
80/0  & Upper bound      & - &  91.23 $\pm$ 0.44	& 6.08 $\pm$ 1.84    \\ [0.1em]

\hline
8/0   & Lower bound      & - & 76.07 $\pm$ 5.02	& 28.75 $\pm$ 0.72    \\
& MT \citep{tarvainen2017mean}     & - & 78.22 $\pm$ 6.89	& 16.74 $\pm$ 4.80	\\
& SASSnet \citep{li2020shape}      & - & 83.70 $\pm$ 1.48	& 16.90 $\pm$ 1.35	\\
& DTC \citep{luo2021semi}          & - & 83.10 $\pm$ 0.26	& \ul{12.62 $\pm$ 1.44}	\\

\cline{2-5}
& UAMT \citep{yu2019uncertainty}   & 8 & \ul{85.09 $\pm$ 1.42}	& 18.34 $\pm$ 2.80	\\
& DUMT \citep{wang2020double}      & 16 & 82.97 $\pm$ 1.76	& 14.43 $\pm$ 0.67  \\
& URPC \citep{luo2022semi}    & 1 & 84.47 $\pm$ 0.31	& 17.11 $\pm$ 0.60	\\

\multirow{-8}{*}{\specialcell{8/72 \\ (10\%)}} &
Ours        & 1 & $\mathbf{86.58 \pm 1.03}^*$	& $\mathbf{11.82 \pm 1.42}$	\\


\hline
\noalign{\vskip\doublerulesep\vskip-\arrayrulewidth} \hline
16/0  & Lower bound     & - & 81.46 $\pm$ 2.96	& 23.61 $\pm$ 4.94  \\
& MT \citep{tarvainen2017mean}     & - & 86.06 $\pm$ 0.81	& 11.63 $\pm$ 3.40  \\
& SASSnet \citep{li2020shape}      & - & 87.81 $\pm$ 1.45	& \ul{10.18 $\pm$ 0.55}	\\
& DTC \citep{luo2021semi}          & - & 87.35 $\pm$ 1.26	& 10.25 $\pm$ 2.49	\\

\cline{2-5}
& UAMT \citep{yu2019uncertainty}   & 8 & 87.78 $\pm$ 1.03	& 11.10 $\pm$ 1.91  \\
& DUMT \citep{wang2020double}      & 16 & 87.42 $\pm$ 0.97	& 10.78 $\pm$ 2.26  \\
& URPC \citep{luo2022semi}    & 1 &  \ul{88.58 $\pm$ 0.10}	& 13.10 $\pm$ 0.60	\\

\multirow{-8}{*}{\specialcell{16/64 \\ (20\%)}} &
Ours        & 1 & $\mathbf{88.60 \pm 0.82}$	& $\mathbf{7.61 \pm 0.78}^*$   \\
\end{tabular}
\label{table:results_LA}
\end{table*}

\subsection{Comparison with the state-of-the-art}
We first compare our method with relevant semi-supervised segmentation approaches and report the quantitative results in Tables~\ref{table:results_LA} and \ref{table:results_FLARE}. The upper and lower bound from the backbone architecture V-Net \citep{milletari2016v} are reported at the top of each section. Furthermore, non-uncertainty-based methods such as MT \citep{tarvainen2017mean}, DTC \citep{luo2021semi}, and SASSnet \citep{li2020shape} and uncertainty-based methods UAMT \citep{yu2019uncertainty}, DUMT \citep{wang2020double}, and URPC \citep{luo2022semi} are included in our evaluation.

\paragraph{(a) Left Atrium segmentation}
Table~\ref{table:results_LA} shows the segmentation performance on the Left Atrium (LA) test set under the standard 10\% (top) and 20\% (bottom) annotation settings. From the top half of the table, we observe that leveraging unlabeled data improves the lower bound across all models. The uncertainty-based approaches typically outperform their non-uncertainty counterparts in terms of DSC, but yield inferior results in terms of HD. Among these methods, UAMT and DTC achieve the best DSC and HD scores, respectively. Nevertheless, compared to these best-performing baselines, our method brings improvements in both DSC (1.5\%) and HD (0.8mm) scores. Moreover, uncertainty estimation in our method requires a single inference from an anatomically-aware representation, whereas UAMT uses $K$=8 inferences per training step to obtain an uncertainty map. This highlights the efficiency of the proposed approach, which yields a better segmentation performance yet requires substantially less computational time at each training step.

Furthermore, we validate our method on the 20\% annotation scenario, whose results are reported in bottom half of Table~\ref{table:results_LA}. We observe a similar trend in these results, with uncertainty-based approaches outperforming non-uncertainty-based methods in DSC, whereas their performance in terms of HD is degraded. An interesting observation is that existing methods are ranked differently across the two annotation settings, indicating that they might be sensitive to the annotation scenario. For example, while UAMT achieves the best DSC score under the 10\% annotation setting, URPC yields the best results in the 20\% annotation case. Similarly, the best models are different for HD metric, i.e., DTC under the 10\% setting and SASSNet in the 20\% setting. In contrast, our method consistently outperforms each existing approach in both DSC and HD scores, highlighting its robustness against the amount of labeled data.

\begin{table*}[!ht]
\centering
\footnotesize
\setlength{\tabcolsep}{3pt}
\caption{\textbf{Segmentation results on the FLARE test set for the 10\% and 20\% annotation settings.} Uncertainty-based methods with $K$ inferences per training step are grouped at the bottom of each section, while $K$ = - indicates non-uncertainty-based methods. Our method produces the best results on average. The best and second-best results are highlighted in bold and underlined, whereas the statistical significance between the top two results is denoted in $^*$. The number of labeled and unlabeled data indicated with $N$ and $M$, respectively.}
\begin{tabular}{s | c | l | c | c c c c c}
\rowcolor{gray!10} & \bf $N$/$M$  & \textbf{Methods}         & \bf \#$K$ & \textbf{Average} & \cellcolor{blue!10} \textbf{Liver} & \cellcolor{green!10} \textbf{Kidney} & \cellcolor{red!10} \textbf{Spleen} & \cellcolor{yellow!10} \textbf{Pancreas} \\
\toprule
& 260/0 & Upper bound      & -     & 85.80 $\pm$ 1.42	& 94.95 $\pm$ 0.30	& 93.20 $\pm$ 0.81	& 89.65 $\pm$ 2.91	& 65.38 $\pm$ 2.57    \\ [0.2em]

\cline{2-9}
& 26/0 &   Lower bound      & -    & 70.09 $\pm$ 2.77	& 88.37 $\pm$ 2.31	& 81.12 $\pm$ 2.49	& 70.74 $\pm$ 4.41	& 40.14 $\pm$ 3.84    \\
& &   MT \citep{tarvainen2017mean}     & -    & 70.76 $\pm$ 2.79	& 88.77 $\pm$ 3.11	& 83.34 $\pm$ 1.22	& 72.91 $\pm$ 4.35	& 38.01 $\pm$ 2.62	\\
& &   SASSnet \citep{li2020shape}      & -    & 61.43 $\pm$ 14.3	& 86.94 $\pm$ 2.88	& 63.59 $\pm$ 43.0	& 59.83 $\pm$ 18.6	& 35.36 $\pm$ 5.05	\\
& &   DTC \citep{luo2021semi}          & -    & 68.07 $\pm$ 1.42	& 87.99 $\pm$ 1.79	& 83.11 $\pm$ 3.93	& 66.04 $\pm$ 3.40	& 35.15 $\pm$ 1.26	\\

\cline{3-9}
& &   UAMT \citep{yu2019uncertainty}   & 8    & \ul{73.63 $\pm$ 0.65}	& $\mathbf{91.65 \pm 0.49}$	& 84.70 $\pm$ 2.39	& \ul{76.16 $\pm$ 2.58}	& \ul{42.01 $\pm$ 2.24}	\\
& &   DUMT \citep{wang2020double}      & 16   & 69.04 $\pm$ 1.39	& 87.28 $\pm$ 0.82	& 80.47 $\pm$ 3.88	& 68.23 $\pm$ 6.79	& 40.18 $\pm$ 2.59  \\
& &   URPC \citep{luo2022semi}        & 1     & 73.31 $\pm$ 1.11	& \ul{91.09 $\pm$ 0.62}	& \ul{85.88 $\pm$ 1.82}	& 75.40 $\pm$ 2.64	& 40.89 $\pm$ 4.05	\\

\multirow{-9}{*}{\rotatebox[origin=c]{90}{\bf DSC (\%) $\uparrow$}} 
& \multirow{-8}{*}{\specialcell{\scriptsize{26/234} \\ (10\%)}}
&   Ours        & 1 & $\mathbf{75.28 \pm 1.54}^*$	& 90.78 $\pm$ 1.26	& $\mathbf{87.09 \pm 1.89}$	& $\mathbf{78.13 \pm 1.23}$	& $\mathbf{45.12 \pm 2.20}^*$	\\ [0.2em]

\hline
& 260/0     & Upper bound           & -   & 6.37 $\pm$ 1.15	& 5.50 $\pm$ 2.86	& 3.31 $\pm$ 1.10	& 7.49 $\pm$ 1.94	& 9.17 $\pm$ 0.66    \\ [0.2em]

\cline{2-9}
& 26/0 &   Lower bound      & - & 18.51 $\pm$ 4.01	& 15.26 $\pm$ 0.90	& 9.89 $\pm$ 2.13	& 30.51 $\pm$ 11.9	& 18.40 $\pm$ 3.53    \\

& &   MT \citep{tarvainen2017mean}    & - & 18.58 $\pm$ 1.66	& 12.09 $\pm$ 3.72	& 8.70 $\pm$ 0.85	& 35.89 $\pm$ 7.47	& 17.64 $\pm$ 1.53	\\
& &   SASSnet \citep{li2020shape}     & - & 27.76 $\pm$ 8.51	& 24.59 $\pm$ 23.0	& 15.1 $\pm$ 11.1	& 51.86 $\pm$ 21.3	& 19.53 $\pm$ 0.89	\\
& &   DTC \citep{luo2021semi}         & - & 23.11 $\pm$ 6.01	& 21.63 $\pm$ 16.7	& 18.8 $\pm$ 11.3	& 32.64 $\pm$ 16.8	& 19.31 $\pm$ 2.07	\\

\cline{3-9}
& &   UAMT \citep{yu2019uncertainty}  & 8 & 14.30 $\pm$ 1.94	& $\mathbf{10.44 \pm 1.45}$	& \ul{8.08 $\pm$ 1.41}	& \ul{20.44 $\pm$ 6.18}	& 18.24 $\pm$ 3.04	\\
& &   DUMT \citep{wang2020double}     & 16 & 22.35 $\pm$ 3.82	& 13.23 $\pm$ 2.28	& 19.21 $\pm$ 13.9	& 36.17 $\pm$ 15.5	& 20.77 $\pm$ 3.58  \\
& &   URPC \citep{luo2022semi}        & 1 & \ul{14.23 $\pm$ 1.97}	& 11.71 $\pm$ 2.37	& $\mathbf{7.41 \pm 1.16}$	& 20.82 $\pm$ 5.02	& \ul{16.96 $\pm$ 3.00}	\\

\multirow{-9}{*}{\rotatebox[origin=c]{90}{\bf HD (mm) $\downarrow$}}
& \multirow{-8}{*}{\specialcell{\scriptsize{26/234} \\ (10\%)}}
&   Ours        & 1 & $\mathbf{13.69 \pm 0.68}$	& \ul{10.85 $\pm$ 1.69}	& 9.48 $\pm$ 2.10	& $\mathbf{18.45 \pm 4.17}$	& $\mathbf{15.98 \pm 1.33}$	\\ [0.2em]

\hline
\hline

& 52/0 &   Lower bound      & - & 70.15 $\pm$ 1.58	& 88.40 $\pm$ 1.24	& 81.91 $\pm$ 2.07	& 68.40 $\pm$ 5.68	& 41.88 $\pm$ 7.44    \\
& &   MT \citep{tarvainen2017mean}     & -    & 72.10 $\pm$ 1.84	& 89.82 $\pm$ 2.30	& 85.15 $\pm$ 1.66	& 71.87 $\pm$ 4.28	& 41.55 $\pm$ 2.99	\\
& &   SASSnet \citep{li2020shape}      & -    & 69.74 $\pm$ 4.43	& 88.41 $\pm$ 1.10	& 86.19 $\pm$ 3.13	& 64.11 $\pm$ 12.1	& 40.25 $\pm$ 3.07	\\
& &   DTC \citep{luo2021semi}          & -    & 68.49 $\pm$ 1.30	& 89.61 $\pm$ 0.71	& 83.31 $\pm$ 4.39	& 62.76 $\pm$ 5.64	& 38.29 $\pm$ 3.38	\\

\cline{3-9}
& &   UAMT \citep{yu2019uncertainty}   & 8    & \ul{74.72 $\pm$ 1.15}	& 89.54 $\pm$ 3.10	& \ul{87.92 $\pm$ 1.52}	& \ul{73.07 $\pm$ 3.91}	& $\mathbf{48.34 \pm 1.41}$	\\
& &   DUMT \citep{wang2020double}      & 16   & 72.08 $\pm$ 2.77	& 90.11 $\pm$ 1.66	& 85.43 $\pm$ 4.82	& 71.83 $\pm$ 0.92	& 40.94 $\pm$ 4.17  \\
& &   URPC \citep{luo2022semi}         & 1    & 74.26 $\pm$ 1.02	& \ul{91.02 $\pm$ 0.54}	& 87.91 $\pm$ 2.47	& 72.06 $\pm$ 1.82	& 46.03 $\pm$ 0.40	\\

\multirow{-9}{*}{\rotatebox[origin=c]{90}{\bf DSC (\%) $\uparrow$}}
& \multirow{-8}{*}{\specialcell{\scriptsize{52/208}  \\ (20\%)}}
&    Ours        & 1 & $\mathbf{76.69 \pm 0.81}^*$	& $\mathbf{91.84 \pm 1.00}^*$	& $\mathbf{88.72 \pm 0.74}$	& $\mathbf{78.07 \pm 0.69}^*$	& \ul{48.14 $\pm$ 1.73}	\\ [0.2em]
\hline

& 52/0 &   Lower bound      & - & 15.63 $\pm$ 0.33	& 15.18 $\pm$ 4.46	& 11.93 $\pm$ 4.64	& 20.50 $\pm$ 2.56	& $\mathbf{14.91 \pm 2.78}$    \\

& &   MT \citep{tarvainen2017mean}     & - & 16.39 $\pm$ 3.34	& $\mathbf{11.04 \pm 0.58}$	& 10.89 $\pm$ 0.91	& 25.70 $\pm$ 9.08	& 17.94 $\pm$ 4.50	\\
& &   SASSnet \citep{li2020shape}      & - & 23.84 $\pm$ 0.79	& 34.01 $\pm$ 14.3	& 11.89 $\pm$ 8.66	& 32.28 $\pm$ 1.53	& 17.16 $\pm$ 1.69	\\
& &   DTC \citep{luo2021semi}          & - & 22.46 $\pm$ 2.12	& 25.23 $\pm$ 20.1	& 18.09 $\pm$ 8.14	& 29.05 $\pm$ 4.84	& 17.46 $\pm$ 1.02	\\

\cline{3-9}
& &   UAMT \citep{yu2019uncertainty}   & 8 & 14.50 $\pm$ 2.46	& 16.60 $\pm$ 4.11	& \ul{7.83 $\pm$ 0.76}	& \ul{17.91 $\pm$ 8.34}	& \ul{15.66 $\pm$ 0.76}	\\
& &   DUMT \citep{wang2020double}      & 16 & 15.53 $\pm$ 2.75	& 11.74 $\pm$ 2.27	& 8.64 $\pm$ 0.95	& 25.43 $\pm$ 8.42	& 16.31 $\pm$ 0.89  \\
& &   URPC \citep{luo2022semi}    & 1 & \ul{14.16 $\pm$ 0.68}	& \ul{11.16 $\pm$ 2.09}	& 8.47 $\pm$ 2.79	& 20.66 $\pm$ 0.80	& 16.33 $\pm$ 1.70	\\

\multirow{-9}{*}{\rotatebox[origin=c]{90}{\bf HD (mm) $\downarrow$}}
& \multirow{-8}{*}{\specialcell{\scriptsize{52/208}  \\ (20\%)}}
&   Ours        & 1 & $\mathbf{13.11 \pm 0.45}$	& 11.32 $\pm$ 2.29	& $\mathbf{7.79 \pm 2.69}$	& $\mathbf{17.38 \pm 4.19}$	& 15.94 $\pm$ 0.28	\\
\end{tabular}
\label{table:results_FLARE}
\end{table*}

\paragraph{(b) Abdominal multi-organ segmentations}
Table~\ref{table:results_FLARE} presents the performance of the abdominal multi-organ segmentations on the FLARE test set. The results of 10\% and 20\% annotation experiments are grouped in the top and bottom half of the table, respectively. We report individual organs as well as average results. From the top half of the table, we first notice that the performance of most existing methods is improved when compared to the lower bound in both DSC and HD scores, except SASSNet, DTC, and DUMT. The gap in the segmentation performance of SASSNet and DTC is due to the use of signed distance maps (SDM), which are designed for binary segmentation. Adopting these methods for multi-class segmentation is challenging since it requires careful hyperparameter tuning of per-class SDM predictions, which is beyond the scope of this work. Note that DUMT did not outperform the simple baseline under a multi-class setting, which is consistent with the observations in \citep{van2022pitfalls}. Among the existing methods, the uncertainty-based methods (UAMT and URPC) perform well in both segmentation metrics. These methods improve the segmentation of liver and spleen regions, achieving the best average DSC and HD scores in UAMT and URPC, respectively. Compared to these best-performing baselines, our method predominantly improves the segmentation of challenging regions, notably the pancreas organ. Overall our anatomically-aware method consistently performs well in all regions and improves average DSC (1.65\%) and HD (0.6mm) scores.

The results of the 20\% annotation scenario are reported in the bottom half of Table~\ref{table:results_FLARE}. We notice a similar trend in the results when compared to the 10\% annotation setting. All existing methods, except SASSNet and DTC, improve the segmentation performance over the lower bound in both DSC and HD scores. Our method outperforms the best-performing baselines (UAMT and URPC) in most cases and improves the average DSC (1.95\%) and average HD (1mm) scores. These results show that our method consistently outperforms the existing approaches across different datasets and labeling scenarios. We can, therefore, argue that including our novel anatomically-aware module is a valuable alternative to existing semi-supervised segmentation approaches.

\begin{figure*}[!ht]
\centering
\includegraphics[width=0.975\linewidth]{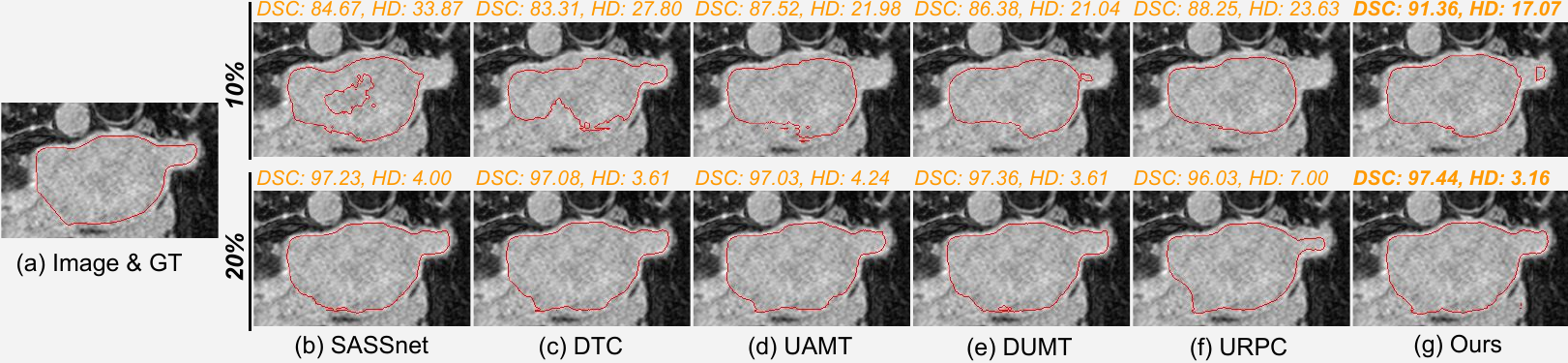}
\caption{\textbf{Qualitative comparison under the 10\% and 20\% annotation settings on LA dataset.} DSC (\%) and HD (mm) scores are mentioned at the top of each image. Each image is overlaid with a contour of segmentation prediction or ground truth (red).}
\label{fig:seg_results_LA}
\end{figure*}

\begin{figure*}[!ht]
\centering
\includegraphics[width=0.975\linewidth]{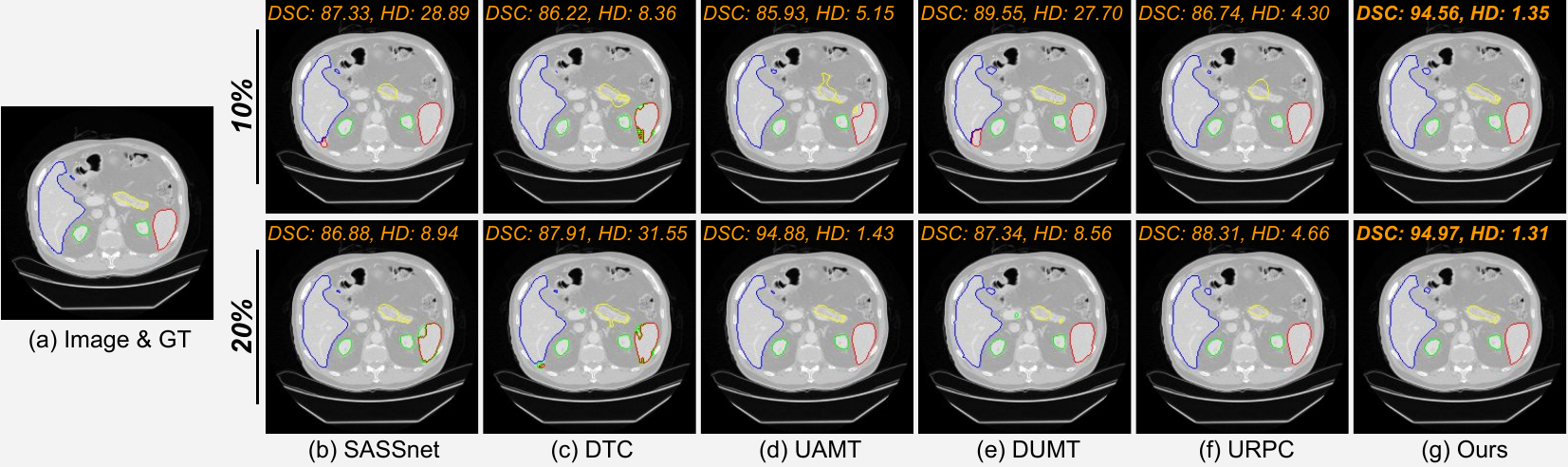}
\caption{\textbf{Qualitative comparison under the 10\% and 20\% annotation settings on FLARE dataset.} Average DSC (\%) and average HD (mm) scores are mentioned at the top of each image.
The colorings are liver (blue), kidney (green), spleen (red), and pancreas (yellow).}
\label{fig:seg_results_FLARE}
\end{figure*}

\subsection{Qualitative Analysis}
Visual results of the left atrium (LA) segmentation obtained by different methods are depicted in Fig.~\ref{fig:seg_results_LA}. In the top row (10\% annotation setting), the existing approaches produce segmentation output with holes (SASSnet, UAMT) and noisy boundaries (SASSnet, DTC, UAMT, DUMT). In contrast, URPC and our methods produce smoother segmentations, but URPC generates under-segmented output compared to our method. Note that a post-processing tool is commonly employed in SASSNet to improve the segmentation performance. However, this is avoided in our experiments for a fair comparison. In the 20\% annotation setting (bottom row), with access to more labeled data, all methods reduce segmentation errors. Even in this case, our method produces promising and smoother segmentations when compared to existing approaches.

To highlight the deficiencies of these approaches in multi-class segmentation, we now show qualitative results on abdominal organs in Fig.~\ref{fig:seg_results_FLARE}. In the 10\% annotation setting (top row), we first observe that misclassification between different organs is a common problem across existing approaches, notably in SASSnet, DTC, UAMT, and DUMT. For instance, part of the liver is segmented as a spleen in SASSnet and DUMT, whereas the parts of the spleen are misclassified as kidneys in DTC and as pancreas in UAMT. This misclassification could be due to either similar intensity characteristics across different organs \citep{durieux2018abdominal} or the inefficiency of networks in discriminating multi-class distributions \citep{van2022pitfalls}. Furthermore, most methods (SASSnet, DTC, UAMT, URPC) have failed to capture the challenging pancreas region. In contrast, our method provides an improved segmentation in this challenging region and minimizes classification errors. In the bottom row of Fig.~\ref{fig:seg_results_FLARE}, adding more labeled images to the training (20\% annotation setting) also reduces classification errors (UAMT, URPC). Our method similarly improves the segmentation performance in all observed regions. The quantitative results from the previous section further support the superiority of our approach. Overall, we argue that the observed improvements in both datasets could be attributed to the knowledge derived from the anatomically-aware representation.


\begin{figure}[!ht]
\centering
\begin{subfigure}{0.232\textwidth}
\centering
\includegraphics[width=1\linewidth,clip]{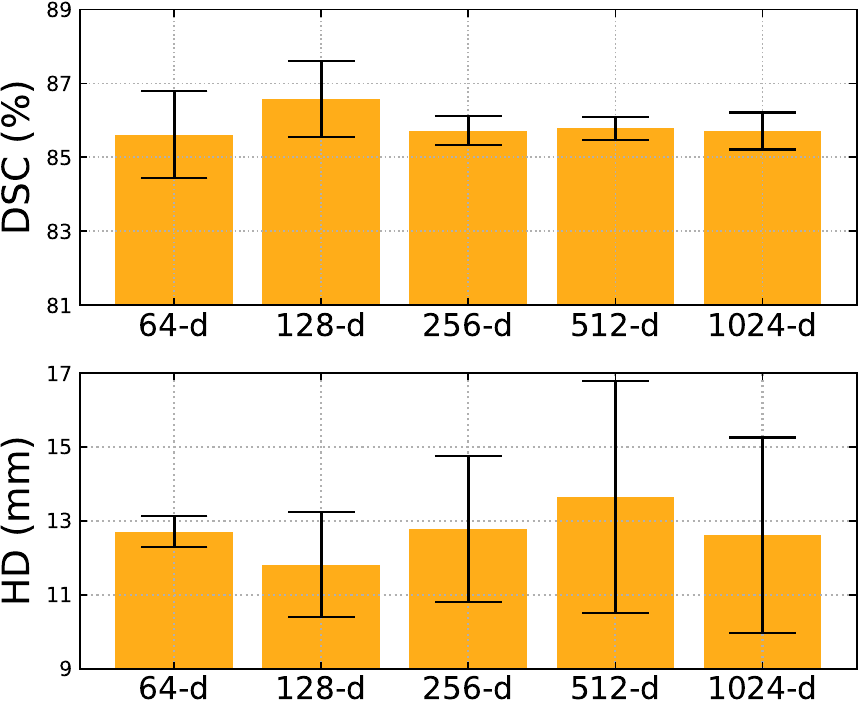}
    \caption{LA dataset}
\end{subfigure}
\begin{subfigure}{0.232\textwidth}
\centering
\includegraphics[width=1\linewidth,clip]{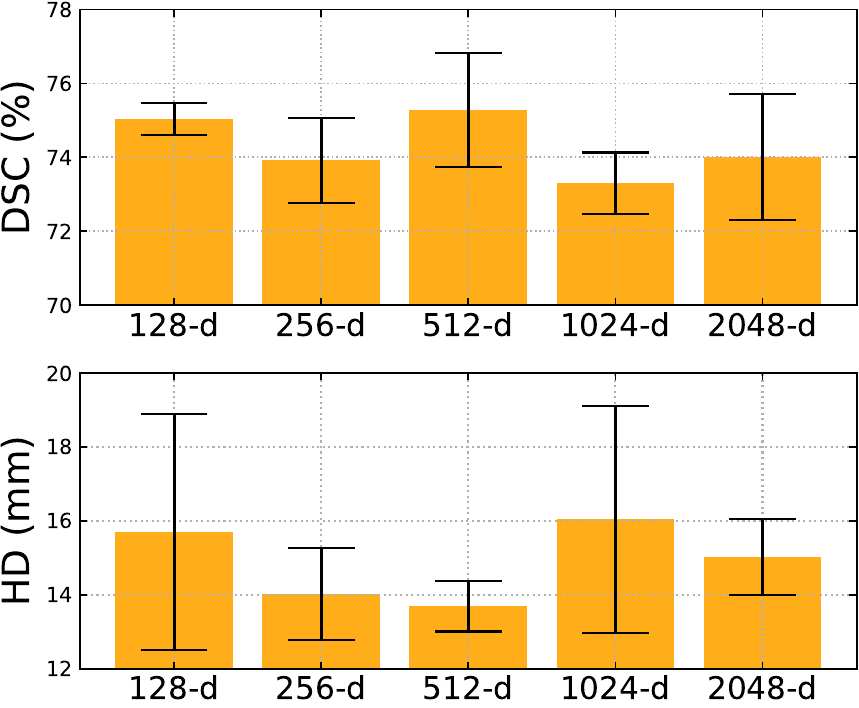}
    \caption{FLARE dataset}
\end{subfigure}
\caption{\textbf{Segmentation performance with different latent space sizes of DAE} - Each bar indicates the DSC (top) and HD (bottom) scores under the 10\% annotation setting. The best results are obtained for the latent space size $d$=128 in binary LA segmentations (a), whereas $d$=512 is needed for abdominal multi-organ segmentations (b).}
\label{fig:LS_dim_study}
\end{figure}

\begin{figure}[!ht]
\centering
\includegraphics[width=0.75\linewidth]{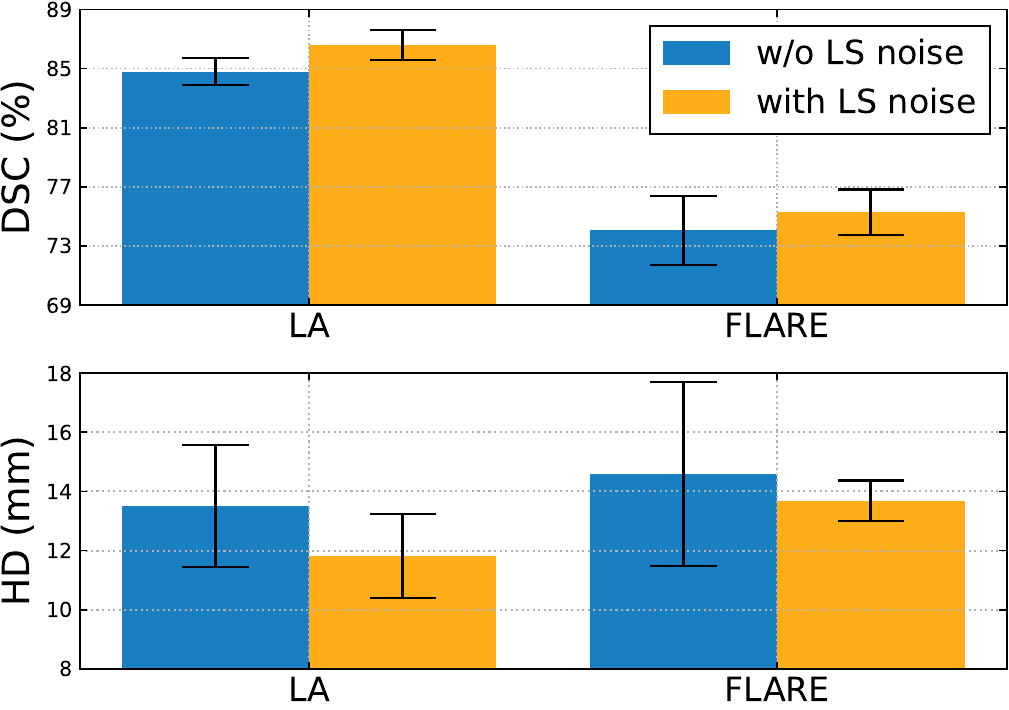}
\caption{\textbf{Impact of noise in the latent space of DAE on segmentation performance} - Each bar indicates the DSC (top) and HD (bottom) scores under the 10\% annotation setting. Addition of a noise (orange) in latent space improves DSC and HD scores.}
\label{fig:LS_noise_study}
\end{figure}

\subsection{Choice of Latent Space in DAE}
\label{sec:choice_of_LS}
Our anatomically-aware prior (DAE) plays a vital role in guiding the segmentation model. Therefore, we investigate the impact of the design choices made in the DAE on the final segmentation performance. The latent space (LS) of our DAE is first studied under varying sizes ($d$) across two datasets in Fig.~\ref{fig:LS_dim_study}. The results show that the segmentation performance varies with LS sizes. The best results are achieved for $d$=128 in binary left atrium segmentations and $d$=512 in abdominal multi-organ segmentations. It indicates that the choice of the latent space size, $d$, depends on the complexity of the dataset.

Furthermore, the LS of the DAE is perturbed with an addition of a Gaussian noise. This facilitates a different set of reconstructions from the DAE when training the segmentation model. The different reconstructions aid in better guiding the segmentation model. To validate this notion, we conduct experiments with and without adding a noise in the LS across both datasets in Fig.\ref{fig:LS_noise_study}. The results demonstrate that the final segmentation performance improves up to 1.79\% in DSC and 1.69mm in HD by adding a noise in the LS of the DAE module. These analyses show the impact of our design choices in the anatomically-aware prior on the segmentation performance.

\begin{table*}[!t]
\footnotesize
\centering
\addtolength{\tabcolsep}{2pt}
\caption{\textbf{Effectiveness of our proposed uncertainty estimation on segmentation results using different strategies.} The number of labeled and unlabeled data indicated with $N$ and $M$, respectively.}
\begin{tabular}{c | l | c c | c c}
\rowcolor{gray!10} & & \multicolumn{2}{c|}{\textbf{LA Dataset}} & \multicolumn{2}{c}{\textbf{FLARE Dataset}} \\
\hline
\rowcolor{gray!10} \bf $N$/$M$ & \textbf{Methods}          & \textbf{DSC (\%)} $\uparrow$ & \textbf{HD (mm)} $\downarrow$ & \textbf{DSC (\%)} $\uparrow$ & \textbf{HD (mm)} $\downarrow$ \\

\toprule
& UAMT \citep{yu2019uncertainty}  & 85.09 $\pm$ 1.42    & 18.34 $\pm$ 2.80      & 73.63 $\pm$ 0.65   & 14.30 $\pm$ 1.94 \\
& Ours (Threshold)                & 85.39 $\pm$ 0.91    & 12.96 $\pm$ 3.05      & 74.25 $\pm$ 1.76    & 14.47 $\pm$ 1.63 \\
& Ours (Entropy)                  & 85.92 $\pm$ 1.52    & $\mathbf{11.16 \pm 0.82}$  & 74.01 $\pm$ 0.62    & 15.03 $\pm$ 2.00 \\
\multirow{-4}{*}{\specialcell{8/72 \\ (10\%)}} &
Ours                            & $\mathbf{86.58 \pm 1.03}$	& 11.82 $\pm$ 1.42       & $\mathbf{75.28 \pm 1.54}$  & $\mathbf{13.69 \pm 0.68}$ \\ [0.2em]
\hline

& UAMT \citep{yu2019uncertainty}  & 87.78 $\pm$ 1.03	& 11.10 $\pm$ 1.91      & 74.72 $\pm$ 1.15   & 14.50 $\pm$ 2.46 \\
& Ours (Threshold)                & 88.12 $\pm$ 1.16    & 8.44 $\pm$ 1.96       & 74.80 $\pm$ 0.80   & 14.09 $\pm$ 1.83 \\
& Ours (Entropy)                  & 87.76 $\pm$ 0.36    & 8.90 $\pm$ 0.48       & 74.57 $\pm$ 0.53   & 15.38 $\pm$ 2.57\\
\multirow{-4}{*}{\specialcell{16/64 \\ (20\%)}} &
Ours                            & $\mathbf{88.60 \pm 0.82}$	& $\mathbf{7.61 \pm 0.78}$    & $\mathbf{76.69 \pm 0.81}$  & $\mathbf{13.11 \pm 0.45}$ \\

\end{tabular}
\label{table:ablation_uncertainty}
\end{table*}

\subsection{Ablation Study on uncertainty}
To validate the effectiveness of our uncertainty estimation on the segmentation performance, we conducted two experiments by adopting a threshold strategy and a predictive entropy scheme used in UAMT. Specifically, a threshold strategy filters out the most unreliable region from the uncertainty map ($U_i$), defined as $H > U_i$ with a threshold, $H$, set with a ramp-up function, as in UAMT \citep{yu2019uncertainty}. In the entropy experiments, we estimate the uncertainty ($U_i$) using the entropy of the DAE prediction ($\hat{p}_{T_i}$) and then combining it in a consistency loss as in Eq~\ref{eq:consistency_loss}. The results of these ablation experiments on the LA and FLARE datasets under the 10\% and 20\% annotation settings are reported in Table~\ref{table:ablation_uncertainty}. Compared to UAMT, our threshold and entropy experiments improve the segmentation performance in both DSC and HD scores in most cases. At the same time, our proposed uncertainty method (Sec.~\ref{ssec:uncertainty_anatomical_prior}) achieves the best performance in all the settings. These results show the merit of our anatomically-aware uncertainty estimation for guiding the segmentation model.

\begin{figure*}[ht!]
\begin{subfigure}{0.235\textwidth}
\centering
\includegraphics[width=1\linewidth,trim={0.2cm 0cm 1.cm 1.5cm},clip]{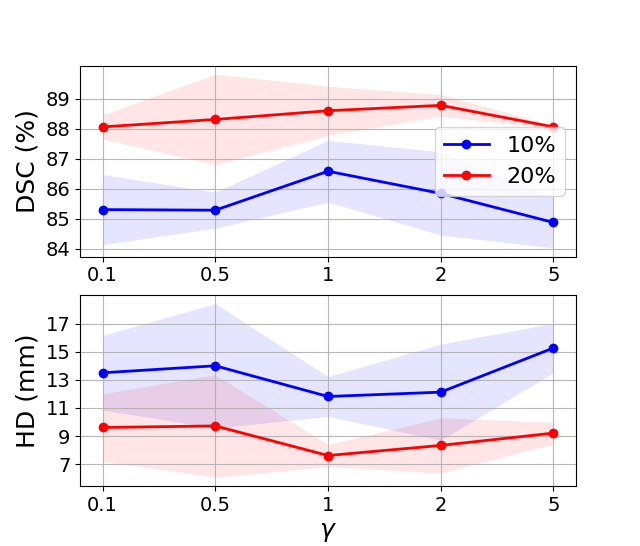}
    \caption{LA - $\gamma$}
    \label{fig:gamma_LA}
\end{subfigure}
\begin{subfigure}{0.235\textwidth}
\centering
\includegraphics[width=1\linewidth,trim={0.2cm 0cm 1.cm 1.5cm},clip]{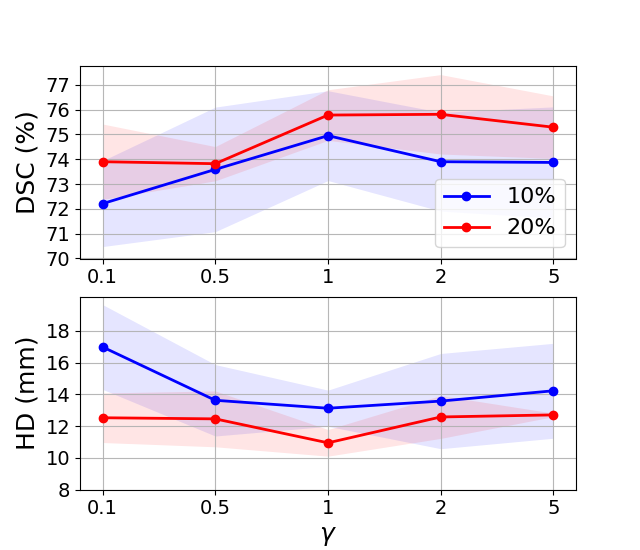}
    \caption{FLARE - $\gamma$}
    \label{fig:gamma_FLARE}
\end{subfigure}
\begin{subfigure}{0.235\textwidth}
\centering
\includegraphics[width=1\linewidth,trim={0.2cm 0cm 1.cm 1.5cm},clip]{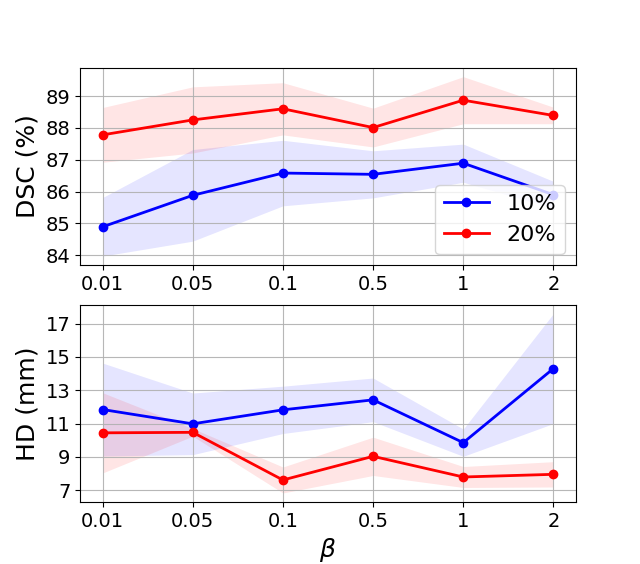}
    \caption{LA - $\beta$}
    \label{fig:beta_LA}
\end{subfigure}
\begin{subfigure}{0.235\textwidth}
\centering
\includegraphics[width=1\linewidth,trim={0.2cm 0cm 1.cm 1.5cm},clip]{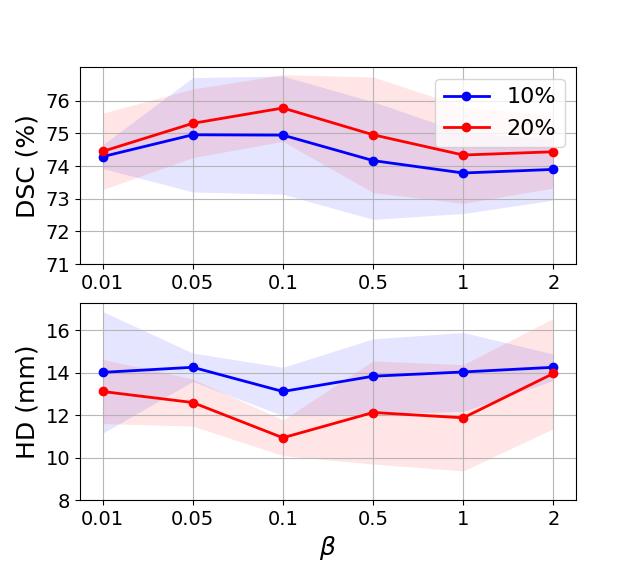}
    \caption{FLARE - $\beta$}
    \label{fig:beta_FLARE}
\end{subfigure}
\centering
\caption{\textbf{Sensitivity of the consistency weight $\beta$ (a, b) and the uncertainty weight $\gamma$ (c, d)} - Each point in a line indicates the DSC (top) and HD (bottom) scores on LA and FLARE datasets under 10\% (blue) and 20\% (red) annotation settings.}
\label{fig:ablation_beta_gamma}
\end{figure*}

\subsection{Impact of \texorpdfstring{$\gamma$}{} and \texorpdfstring{$\beta$}{} hyperparameters}
\label{sec:beta_gamma}
The sensitivity of the uncertainty weight $\gamma$ (in Eq.\ref{eq:consistency_loss}) and the consistency weight $\beta$ on the segmentation performance is shown in Fig.~\ref{fig:ablation_beta_gamma}. In particular, we evaluate the segmentation performance using DSC and HD scores by varying the $\gamma$ and $\beta$ values across the LA and FLARE datasets. In Fig.~\ref{fig:ablation_beta_gamma}(a)-(b), increasing the gamma value leads to an improvement in the segmentation performance in both DSC and HD scores across both datasets. The best results are usually observed for $\gamma = 1$. Beyond that, performance generally decreases, possibly due to an exponential decrease in the weight (Eq.\ref{eq:consistency_loss}) of the reliable target regions.

Figure~\ref{fig:ablation_beta_gamma}(c)-(d) shows the segmentation performance for varying the $\beta$ values. The results show that increasing the beta value improves the segmentation performance. The best result is achieved for $\beta$=0.1 except in the LA dataset (in Fig.~\ref{fig:ablation_beta_gamma}(c)), where $\beta$=1 produces the best scores. Nevertheless, we chose to set $\beta$=0.1 across all our experimental scenarios, as this value is widely adopted in the literature on consistency-based approaches \citep{tarvainen2017mean,wang2021tripled} and for a fair comparison with our baselines \citep{yu2019uncertainty,wang2020double,luo2022semi}.


\begin{table}[!ht]
\footnotesize
\centering
\addtolength{\tabcolsep}{1pt}
\caption{\textbf{Comparison of average training times in seconds per iteration.} Our method adds a minimal overhead on top of the MT approach for uncertainty estimation.}
\begin{tabular}{l | c | c | c}
\bf Methods          & \bf \#$K$ & \bf LA & \bf FLARE \\
\toprule
MT \citep{tarvainen2017mean}    & - & 0.612 & 1.108 \\
SASSnet \citep{li2020shape}     & - & 1.442 & 5.856 \\
DTC \citep{luo2021semi}         & - & 0.989 & 4.874 \\
\hline
UAMT \citep{yu2019uncertainty}  & 8 & 1.207 & 2.429 \\
DUMT \citep{wang2020double}     & 16 & 3.804 & 7.678 \\
URPC \citep{luo2022semi}        & 1 & 0.779 & 1.504 \\
Ours                            & 1 & 0.745 & 1.266 \\
\end{tabular}
\label{table:training_time}
\end{table}

\subsection{Training time}
\label{sec:training_time}
To evaluate the speed of our uncertainty estimation, we compare the computation time required for each training iteration by the proposed and the baseline methods in Table~\ref{table:training_time}. From the table, we observe that the non-uncertainty-based methods (SASSnet, DTC) are slower when compared to uncertainty-based methods across both datasets, LA and FLARE. The relative slow speed of SASSnet and DTC is attributed to the additional computational overhead required for predicting the signed distance maps (SASSnet, DTC) and the inclusion of a discriminator module (SASSnet). On the other hand, ours and the URPC method are faster than the MCDO-based methods (UAMT and DUMT) due to the need of only one inference when estimating the uncertainty (\#$K$=1). Overall, our approach adds a minimal overhead on top of the mean teacher (MT) approach for estimating uncertainty while producing superior segmentation results on both datasets.

\begin{figure*}[!ht]
\centering
\includegraphics[width=0.975\linewidth]{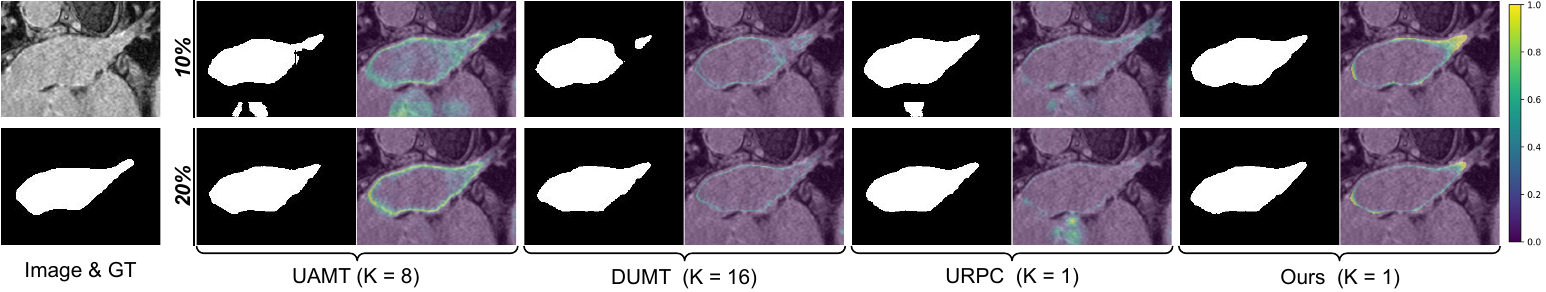}
\caption{\textbf{Uncertainty analysis on the left atrium dataset} - Prediction and uncertainty map (overlaid on its image)  are shown for each uncertainty-based method. The number of inferences for generating the uncertainty map is denoted as $K$.}
\label{fig:uncertainty_pred_maps}
\end{figure*}

\subsection{Uncertainty Analysis}
The predicted segmentation and uncertainty map from different uncertainty-based methods are shown in Fig.~\ref{fig:uncertainty_pred_maps}. The top row shows the 10\% annotation setting, where uncertainties are all over the predicted regions for UAMT. These uncertainties inside the prediction regions are reduced in DUMT, possibly due to more inferences and the addition of feature uncertainty. However, the uncertainties are highly focused on the prediction boundaries. The uncertainty is produced at arbitrary regions in URPC due to their multi-scale discrepancy-based uncertainty estimation. Our method produces uncertainty in challenging regions, such as unclear anatomical boundaries or annotator cuts (as in pulmonary veins), which are estimated using anatomically-aware representation. In the below row of Fig.~\ref{fig:uncertainty_pred_maps}, increasing labeled samples (i.e., 20\% setting) improves the predictions and uncertainty in most cases. Nevertheless, uncertainties are all over the boundaries, or arbitrary regions remain in the existing methods. Our method further improves the uncertainties due to the improvement of anatomically-aware representation using more access to labels. Moreover, our method requires single inference when compared to entropy-based methods.

\section{Discussion and Conclusion} 
\label{sec:discussion_conclusion}
This work proposes a novel anatomically-aware uncertainty estimation method for semi-supervised image segmentation. Our approach consists of leveraging an anatomically-aware representation of labeling masks to estimate the segmentation uncertainty. The obtained uncertainty maps guide the training of the segmentation model within reliable regions of the predicted masks. Our experimental results demonstrate that the proposed method yields improved segmentation results when compared to state-of-the-art baselines on two publicly available benchmarks using left atria and abdominal organs. The qualitative results also show how our anatomically-aware approach improves segmentation in challenging image areas. The ablation studies demonstrate the effectiveness and robustness of our uncertainty estimation when compared to entropy-based methods. Adding noise in the latent space of our representation helps to map the predictions into a better set of plausible segmentations, which improves the segmentation accuracy.
Unlike most uncertainty-based approaches, our anatomically-aware uncertainty requires a single inference, thereby reducing computational complexity.
Moreover, as our anatomically-aware representation is independent of any image information, it can be further enhanced with existing segmentation masks from different datasets or imaging modalities \citep{karani2021test}, potentially further improving the modeling capacity of our representation. The learning representation with an additional constraint can also be explored separately as a post-processing tool that maps the erroneous prediction into anatomically-plausible segmentation \citep{larrazabal2020post,painchaud2020cardiac}. Additionally, our anatomically-aware representation prior could also benefit from the image intensity information to learn a joint representation \citep{oktay2017anatomically,judge2022crisp} for uncertainty estimation in a limited supervision problem. 
Overall, our proposed approach could be leveraged to a broader range of applications where uncertainties could be related to anatomical information.

\section*{Acknowledgements}
This research work was partly funded by the Canada Research Chair on Shape Analysis in Medical Imaging, the Natural Sciences and Engineering Research Council of Canada (NSERC), and the Fonds de Recherche du Quebec (FRQNT). Computational resources have been partially provided by Compute Canada.

\bibliography{biblio}
\end{document}